\documentclass[10pt,twocolumn,letterpaper]{article}

\usepackage{iccv}
\usepackage{times}
\usepackage{epsfig}
\usepackage{graphicx}
\usepackage{amsmath}
\usepackage{amssymb}
\usepackage{multirow}
\usepackage{wrapfig}
\usepackage{bbding}

\usepackage[dvipsnames]{xcolor}

\usepackage[pagebackref=true,breaklinks=true,letterpaper=true,colorlinks,citecolor=ForestGreen, bookmarks=false]{hyperref}

\iccvfinalcopy 

\def\iccvPaperID{****} 

\ificcvfinal\fi

\begin{document}

\title{Multi-Agent Reinforcement Learning Based Frame Sampling for Effective Untrimmed Video Recognition}

\author{Wenhao Wu~$^{1,2}$\thanks{This work was done when Wenhao Wu was a research intern at Baidu.}\quad
Dongliang He~$^{3}$\quad
Xiao Tan~$^{3}$\quad
Shifeng Chen~$^{1}$\thanks{Corresponding author (e-mail: shifeng.chen@siat.ac.cn).}\quad 
Shilei Wen~$^{3}$ \\
$^1$ Shenzhen Institutes of Advanced Technology, Chinese Academy of Sciences, China\\
$^2$ University of Chinese Academy of Sciences,  China \\
$^3$ Department of Computer Vision Technology (VIS), Baidu Inc., China\\
{\tt\small \{wh.wu,shifeng.chen\}@siat.ac.cn ~~~ \{hedongliang01,tanxiao01,wenshilei\}@baidu.com } 
}

\maketitle
\ificcvfinal\thispagestyle{empty}\fi

\begin{abstract}
   Video Recognition has drawn great research interest and great progress has been made. A suitable frame sampling strategy can improve the accuracy and efficiency of recognition. However, mainstream solutions generally adopt hand-crafted frame sampling strategies for recognition. It could degrade the performance, especially in untrimmed videos, due to the variation of frame-level saliency. To this end, we concentrate on improving untrimmed video classification via developing a learning-based frame sampling strategy. We intuitively formulate the frame sampling procedure as multiple parallel Markov decision processes, each of which aims at picking out a frame/clip by gradually adjusting an initial sampling. Then we propose to solve the problems with multi-agent reinforcement learning (MARL). Our MARL framework is composed of a novel RNN-based context-aware observation network which jointly models context information among nearby agents and historical states of a specific agent, a policy network which generates the probability distribution over a predefined action space at each step and a classification network for reward calculation as well as final recognition. Extensive experimental results show that our MARL-based scheme remarkably outperforms hand-crafted strategies with various 2D and 3D baseline methods. Our single RGB model achieves a comparable performance of ActivityNet v1.3 champion submission with multi-modal multi-model fusion and new state-of-the-art results on YouTube Birds and YouTube Cars.
\end{abstract}

\section{Introduction}
Recently, video recognition has attracted great research interest in the computer vision community, due to its importance in real-world applications such as video surveillance, video search, and video recommendation. A video contains a sequence of frames, both spatial information, and temporal relation are important for accurate recognition. 
Compared to recognizing well-trimmed clips, untrimmed videos pose a more critical challenge since not all the frames consistently respond to the specified ground-truth label. Picking out the most informative frames can be an effective method for recognition.

\begin{figure}[t]
\begin{center}
\includegraphics[width=0.9\linewidth]{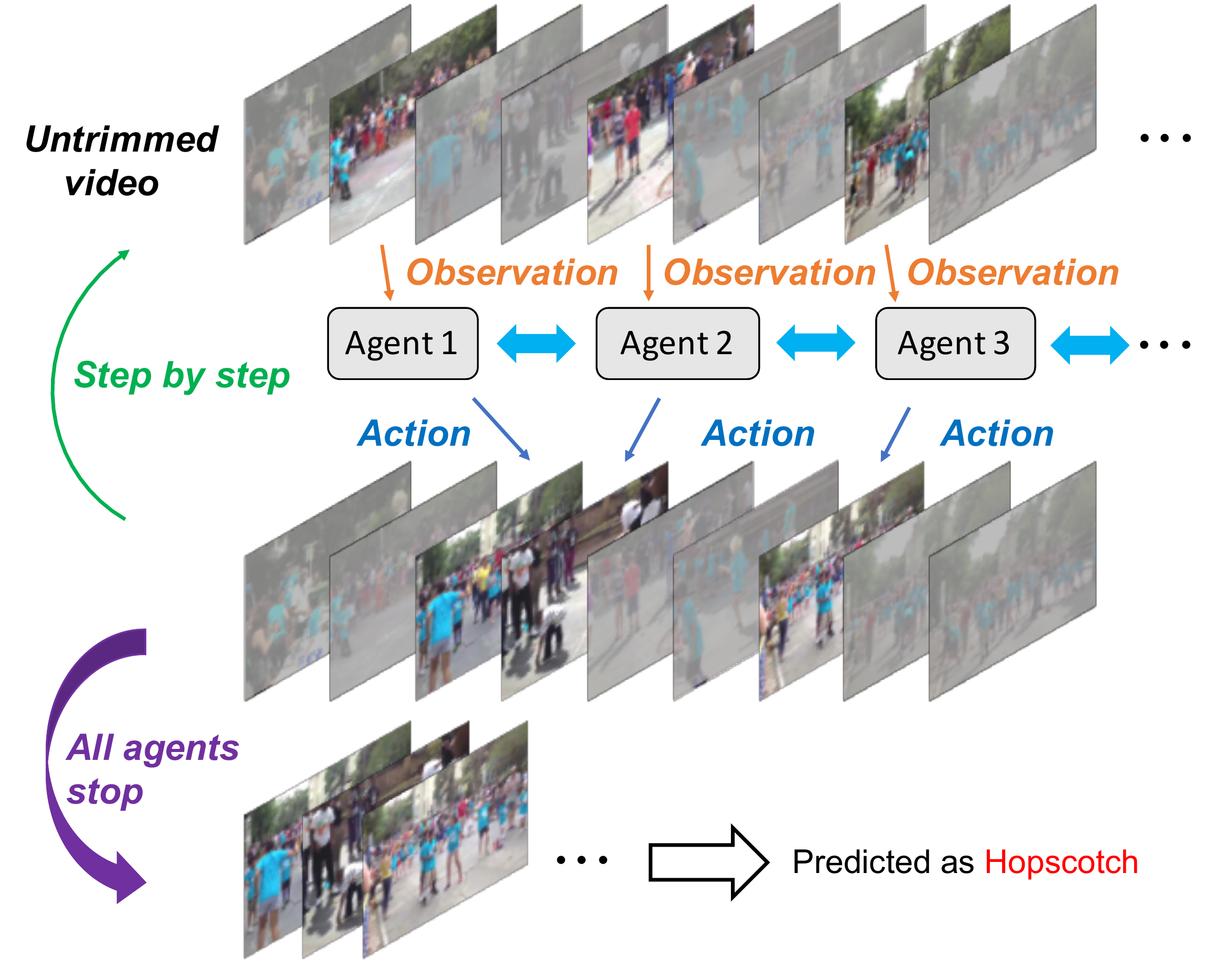}
\end{center}
   \caption{A demonstration of our approach. At each time step, each agent relies on context information among nearby agents to take action for adjusting its sampling location. When all agents stop, the prediction is emitted by the classification decision.}
\label{fig:overview}
\end{figure}

Existing research efforts~\cite{tsn,two-stream,gan2016webly,two-stream-stresnet,gan2016you,tle,T-Resnet,c3d,i3d,s3d,p3d,nonlocal,LRCN,netvlad,actionvlad,attentioncluster,shuttleNet,r2+1d,ARN,eco,mfnet,stnet} mainly focus on building effective and efficient video modeling networks. Generally speaking, they can be divided into two directions, namely, (1) two-stage solutions \cite{LRCN,netvlad,actionvlad,attentioncluster,shuttleNet} which extract spatial feature vectors from video frames and then integrate the obtained local feature sequence into one compact video descriptor for recognition; (2) the 2D  \cite{tsn,two-stream,two-stream-stresnet,tle} or 3D convolution based \cite{c3d,T-Resnet,i3d,s3d,p3d,nonlocal,r2+1d,ARN,eco,mfnet} end-to-end video classification methods. Though great progress has been achieved by these methods, limited attention is paid to the aforementioned variation of frame-level salience among different frames. The mainstream methods all propose to sample input frames by hand-crafted strategies, for instance, evenly sampling $N$ frames/clips from the original video is done in \cite{tsn,stnet,attentioncluster}, sampling $N$ successive frames from a video is adopted by \cite{c3d,p3d,nonlocal}, and directly feeding all video frames in the test phase is chosen in \cite{i3d,s3d,netvlad,shuttleNet}. Sampling $N$ frames/clips either evenly or successively from an untrimmed video cannot guarantee optimum. Meanwhile, feeding all the frames is a sort of brute-force and introduces a much unnecessary computation burden. Therefore, in this paper, we concentrate on how to pick out the most discriminative frames from untrimmed videos to achieve better classification performance.

Intuitively, an effective algorithm for humans to pick out $N$ representative frames from an untrimmed video can be as follows: we first observe $N$ scenes of the whole video, according to the initial observation, then we infer where to check next time until we find out the satisfying frames round by round. Motivated by this, we proposed to formulate frame selection as $N$ parallel Markov decision processes. As is known, a Markov decision process can be naturally modeled by the reinforcement learning framework \cite{li2017deep}. Inspired by the success of reinforcement learning in solving sequential decision-making problems, we propose a multi-agent reinforcement learning (MARL) framework to select multiple discriminative frames or video clips from an untrimmed video to improve the recognition performance. The workflow of our system is illustrated in Figure \ref{fig:overview}. 

Specifically, there are $N$ agents in our framework and each of them is responsible for selecting one informative frame/clip from an untrimmed video. They initially sample $N$ frames/clips evenly from the entire video and iteratively decide that each of their samples should come from temporally preceding or later location until encountering a STOP action at $T_{stop}$-th step or the maximum step number $T_{max}$ is reached. To both combine nearby context and track previous status information for better decision making, we design a shared RNN-based \emph{context-aware observation network} to model the local environment and its nearby context information as well as historical states to generate a status vector for each agent. Conditioned on the status vector, the \emph{policy network} estimates the probability distribution over the predefined action space, according to which each agent takes action to adjust its sampling location. A carefully designed reward function is proposed and the MARL framework is optimized following REINFORCE \cite{williams1992simple} by maximizing the expected reward.

To verify the effectiveness of our proposed multi-agent reinforcement learning framework for frame sampling, extensive experiments are conducted on several popular untrimmed video classification datasets, including ActivityNet v1.2 and v1.3 \cite{caba2015activitynet}, YouTube Birds and YouTube Cars \cite{zhu2018fine}. Results show that the proposed scheme achieves remarkable improvement over 2D/3D CNN baseline solutions which are equipped with different hand-crafted sampling strategies. In more detail, a new state-of-the-art on YouTube Birds and YouTube Cars is achieved and our single RGB model reaches a comparable performance of ActivityNet v1.3 champion submission with multi-modal multi-model fusion. 
To sum up, we make the following contributions.
\begin{itemize}
\item We focus on a previously overlooked point, i.e., the frame sampling strategy, in improving untrimmed video classification performance and intuitively formulate it as a Markov decision process.
\item Multi-agent reinforcement learning is adopted to solve the formulated sequential decision problems. A novel framework that takes both context information and historical environment states into consideration for decision making is designed. 
\item The proposed method can be effectively applied to various existing untrimmed video recognition models to improve the performance, which is well witnessed by the excellent experimental results. 

\end{itemize}

\section{Related Work}
\subsection{Action Recognition}
Our paper is closely related to works on deep-learning based action recognition, including end-to-end convolutional classification networks and two-stage recognition solutions. Karpathy \etal firstly introduces CNN for video classification in \cite{karpathy2014}. Then two-stream ConvNet \cite{two-stream} is proposed to merge the predicted scores from RGB and optical flow modalities, and the performance is improved by a large margin. ST-ResNet \cite{two-stream-stresnet} further introduces residual connections between the two streams of \cite{two-stream} and shows great advantages in results. Currently, TSN \cite{tsn} and C3D \cite{c3d} are two well-known baseline methods for video recognition. The former is a 2D CNN based approach while the latter is based on 3D CNN. There are numerous follow-up studies to improve the aforementioned two baselines. For example, TLE \cite{tle}, ShuttleNet \cite{shuttleNet}, AttentionClusters \cite{attentioncluster} and NetVlad \cite{netvlad,actionvlad} are proposed for better local feature integration instead of directly AVG-Pooling as used in TSN. OFF \cite{off} and motion feature network \cite{motionfeatnet} are proposed for integrating motion information modeling into a spatial CNN network, instead of using two streams. I3D \cite{i3d} inflates deeper networks than C3D for spatial-temporal modeling. Given the heavy computation overhead of I3D, a series of works \cite{s3d,p3d,mfnet,eco,r2+1d,ARN,stnet} are done to strike good effectiveness-efficiency trade-off. Meanwhile, nonlocal network \cite{nonlocal}, compact generalized nonlocal network \cite{yue2018compact} and Nonlocal + GCN \cite{Wang_2018_ECCV} bring great performance gain by leveraging extra network modules to capture feature-point-wise or ROI-wise spatial-temporal relations.   
The frame sampling strategies of all the aforementioned state-of-the-arts are hand-crafted. Instead, we propose to use the learning-based strategy to improve recognition performance for untrimmed videos.

\begin{figure*}
\begin{center}
\includegraphics[width=\linewidth]{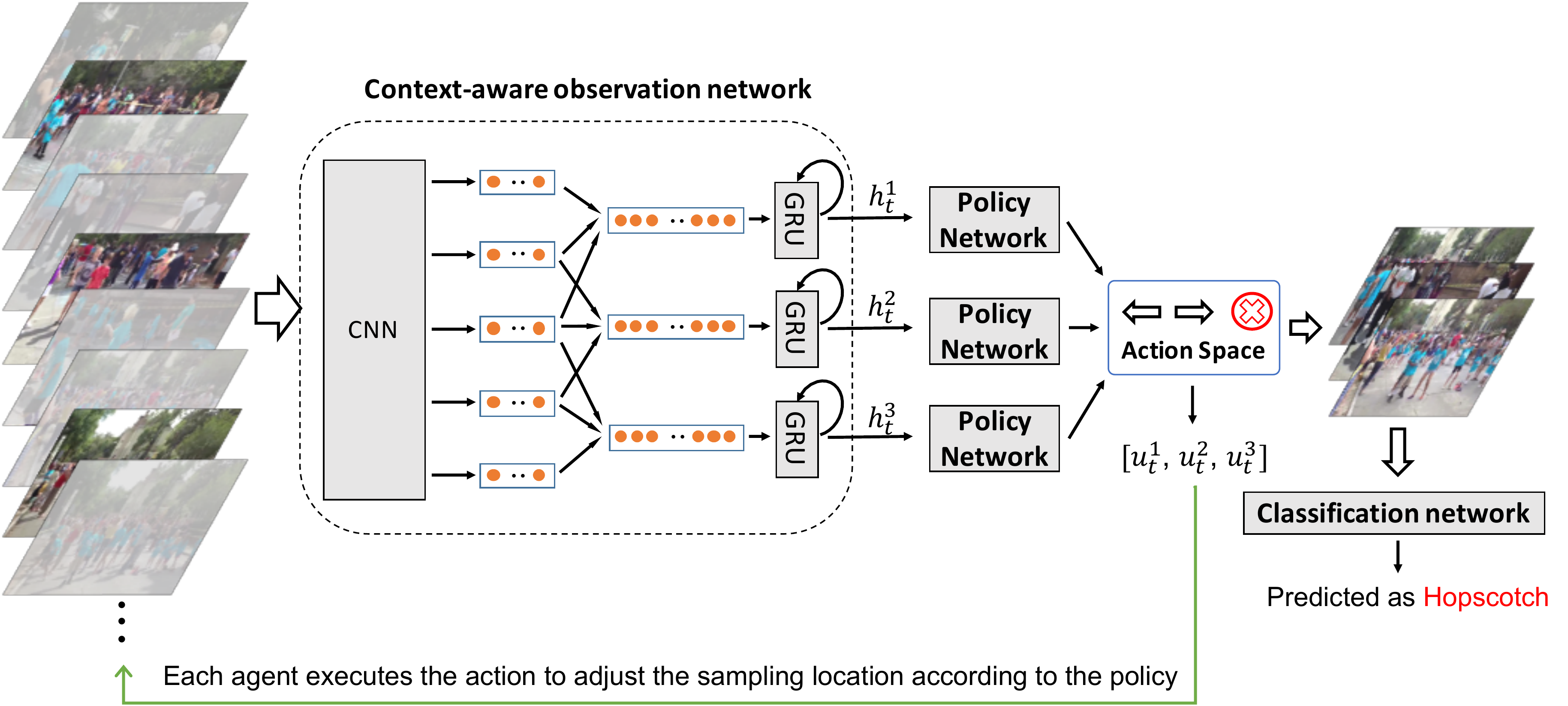}
\end{center}
   \caption{The proposed multi-agent reinforcement learning framework is composed of a context-aware observation network for capturing environment state $h_t^n$, a policy network for estimating probabilistic distribution over action space and a classification network for video-level prediction. There are $N$ (which is 3 for illustration) agents and the environment is the $F$ sampled frames/clips in our system. The agent interacts with the environment by taking action $u_t^n$ to adjust sampling locations iteratively. The context-aware observation network is designed to allow context information communication among nearby agents. GRU, which integrates preceding and current states, is used to model the sequential decision making property of our system.}
\label{fig:MARL}
\end{figure*}

\subsection{Reinforcement Learning}
Our work is largely inspired by those on leveraging reinforcement learning to solve vision problems. \cite{mnih2014recurrent} is a classical work that uses RL for the spatial attention policy in image recognition, as well as in \cite{liu2017localizing}. \cite{Li_2018_CVPR} proposes an aesthetic aware reinforcement learning framework to learn the policy for image cropping and achieves promising results. RL is also adopted for object detection in \cite{Pirinen_2018_CVPR}, for video object segmentation in \cite{han2018reinforcement} and object tracking in \cite{Guo_2018_ECCV,Ren_2018_ECCV}.

As for semantic level video understanding, RL has also played an important role. For instance, it is utilized for action detection in \cite{lifeifei_loc}, for action anticipation in \cite{gao2017red}, for natural language grounding in \cite{he2019read} and for video description in \cite{Wang_2018_CVPR}. Our work is most closely related to the ones that use reinforcement learning for action recognition. In \cite{Chen_2018_ECCV}, the part activation policy of human body parts is learned with RL for action prediction. \cite{tang2018deep} focuses on skeleton-based action recognition and RL is adopted to distinguish discriminative poses. Following \cite{lifeifei_loc}, the Fast-Forward algorithm is proposed in \cite{fan2018watching} to reduce the computation burden for untrimmed video classification. In this work, RL is utilized for both frames skipping planning and early stop decision making. In \cite{wu2018adaframe}, the authors also focus on fast video prediction by adaptively selecting relevant frames on a per-video bias using reinforcement learning. Closely related work mainly focuses on improving prediction efficiency and a single agent is used for decision making. In contrast, we emphasize improving untrimmed video classification baselines and our model adopts multiple agents to cooperatively select multiple frames. 

\section{Approach}
In this paper, we formulate frames selection as multiple sequential decision-making problems. 
Therefore, it naturally fits into the reinforcement learning framework. Figure \ref{fig:MARL} illustrates the multi-agent reinforcement learning framework of our proposed model. 
The model can be seen as multiple agents that interact with a video sequence of $F$ frames/clips over time. Each agent picks a specific
frame out of $F$ frames and employs a \emph{context-aware observation network} to encode the explored environment into a vector, $h_t^a$ which is then fed into the following \emph{policy network} to generate a proper action $u_t^a$ from the action space $\mathcal{A}$. This action adjusts the frame that the agent to pick at the next time step. $N$ agents are identified by $a \in A \equiv \lbrace1,2,...,N\rbrace$. All agents are initialized to be evenly distributed over the temporal space. When all agents decide to pick the current frame (to take the STOP action), a \emph{classification network} emits the prediction based on the selected frames.

\subsection{Architecture}
\textbf{Context-aware observation network} The context-aware observation network is composed of a basic observation network $f_o$ parameterized by $\theta_o$, followed by a context network. The basic observation network is used to encode the video information at the frame/clip selected by the agent (namely, the local environment) into a feature vector $o_t^a$ providing it as input to the context network. CNN based networks, both 2D CNN and 3D CNN are good at capturing features in frames or frame clips, and hence either of them can be adopted as the basic observation network. Unlike the single-agent system, for the multi-agent system, the action selected by each agent not only depends on its local environment state but is also impacted by its context information. Hence, we design a context-aware module $f_h$ upon the basic observation network, parameterized by $\theta_h$, to maintain a joint internal state of agents which summarizes history context information by a recurrent neural network. To make it work effectively, each agent only accesses context information from its $2M$ neighboring agents but not from all agents. More formally, at the time step $t$, the agent $a$ observes a concatenated state $s_t^a = [o_t^{a-M}, ...,o_t^{a}, o_t^{a+M}]$ and its previous hidden states $h_{t-1}^a$ as inputs of the context module, then produces its current hidden states $h_t^a$:
\begin{equation}
    h_t^a = f_h(h_{t-1}^a, s_t^a; \theta_h).
\end{equation}
Especially, if $M$ is set to 1, then the first agent loses the rear information while the last agent loses the front information. We use the beginning and ending frame/clip of the video to compensate for the information needed for the two cases respectively. For the multi-agent model, all of these agents share the context-aware observation network.

\textbf{Policy network} We adopt a fully connected layer followed by a softmax function as the policy network parameterized by $\theta_u$. In every time-step $t$, each agent $a$ selects an action $u_t^a \in U$ to execute, according to the probability distribution $\pi(u_t^a|h_t^a; \theta_{u})$ generated by the policy network. The action space consists of three predefined actions, namely, \emph{moving ahead}, \emph{moving back} and \emph{staying}. The moving stride is set to be $\delta$ frames/clips. When all agents choose \emph{staying}, it means a STOP action is encountered.  Note that we still make parameters of the policy network shared among all agents so that our model is readily applicable for testing at arbitrary $N$. 

\textbf{Classification network} The classification network $f_p$ is parameterized by $\theta_p$. At each time step, these selected frames go through the classification network to produce its corresponding prediction logit $l_t^a\in\mathbb{R}^C$, where $C$ is the number of classes. At the last time step, these prediction scores $l_{T_{stop}} = \lbrace l_{T_{stop}}^1, l_{T_{stop}}^2, ..., l_{T_{stop}}^N \rbrace$ before Softmax normalization are aggregated with average pooling to yield the final video-level prediction. In our proposed method, the classification network can be easily replaced with any video classification module, such as Two Stream \cite{two-stream}, TSN \cite{tsn} and 3D CNNs \cite{c3d}. For the simplicity in design, the classification network shares the parameters of layers before the last classifier layer with the observation network.

\subsection{Objectives}
The overall objective of the MARL-framework is to simultaneously maximize the expected reward of the frame sampling network and minimizing the classification loss. We use standard back-propagation to train the classification network parameterized by $\theta_p$, and REINFORCE \cite{williams1992simple} to optimize the parameters of the basic observation network, the context module and the policy network $\theta_{\pi} = \lbrace \theta_o, \theta_h, \theta_u \rbrace$. Hence, our loss function consists of the MARL loss $\mathcal{L}_{MARL}(\theta_{\pi})$ and the classification loss $\mathcal{L}_{Cls}(\theta_p)$. 

\subsubsection{MARL Objective}
\textbf{Reward function} The reward function reflects how good the actions taken by the agents. When all the agents take actions, each agent gets its own reward $r_t^a$ based on the classification probability $p_t^a = Softmax(l_t^a)$ of the $t$-th time step. The reward is given to encourage the agent to find a more informative frame which can improve the probability of correct prediction step by step. Hence, we design a simple reward function that encourages the model to increase its confidence. Specifically, for the $t$-th ($t>1$) time step, agent $a$ receives a reward follows:
\begin{equation}
    r_t^a = p_{t,gt}^a - p_{t-1,gt}^a,
\end{equation}
where $p_{t,c}^a$ represents the probability of predicting the video as class $c$ at the $t$-th time step for agent $a$, and $gt$ is the ground-truth label of the video. All agents share the same reward function. Considering the sequential decision-making scenario, it is more appropriate to consider a cumulative discounted rewards, where rewards obtained in the more distant future contribute less to the current step. Specifically, at the time step $t$, the discounted return for agent $a$ is 
\begin{equation}
    R_t^a = \sum_{k=0}^{T_{stop}-t} \gamma^k r_{k+t}^a,
\end{equation}

where $\gamma \in (0,1]$ is a constant discount factor that controls the importance of future rewards. 

\textbf{Policy gradient} Given $\mathcal{U}$, a space of action sequences. Following REINFORCE \cite{williams1992simple}, our objective can be expressed as 
\begin{equation}
    J(\theta_{\pi}) = \sum_{a=1}^N\sum_{u\in \mathcal{U}}\pi(u|s;\theta_{\pi})R^a \label{con:expected advantage}.
\end{equation}
In our case we wish to learn network parameters $\theta_{\pi}$ that maximize the equation (\ref{con:expected advantage}). The gradient of $J(\theta_{\pi})$ is
\begin{equation}
    \nabla_{\theta_{\pi}}J(\theta_{\pi}) = \sum_{a=1}^N\sum_{u\in \mathcal{U}}\pi(u|s; \theta_{\pi})\nabla_{\theta_{\pi}}\log \pi (a|s;\theta_{\pi})R^a.
\end{equation}
This leads to a non-trivial optimization problem due to the high dimension of the action sequence space. REINFORCE addresses this by using Monte Carlo sampling to obtain $K$ interaction sequences to approximate the policy gradients:
\begin{equation}
    \nabla_{\theta_{\pi}}J(\theta_{\pi}) \approx \frac{1}{K}\sum_{k=1}^K\sum_{a=1}^N\sum_{t=0}^{T_{stop}}\nabla_{\theta_{\pi}}\log \pi (u_{t,k}^a|s_{t,k}^a;\theta_{\pi})R_t^a.
\end{equation}

Then we can use stochastic gradient descent to minimize the loss function:
\begin{equation}
    \mathcal{L}_J(\theta_{\pi}) = -\frac{1}{K}\sum_{k=1}^K\sum_{a=1}^N\sum_{t=0}^{T_{stop}}\log \pi (u_{t,k}^a|s_{t,k}^a;\theta_{\pi})R_t^a.
\end{equation}

\textbf{Maximum entropy} To prevent policies from becoming deterministic too quickly, researchers use entropy regularization \cite{williams1991function, mnih2016asynchronous}. Its success has sometimes been attributed to the fact that it ``encourages exploration'' \cite{mnih2016asynchronous}. The greater the entropy, the more ability of exploration an agent will have. Therefore, we follow the practice of using the entropy of policy to increase the ability to explore by:
\begin{equation}
    \mathcal{L}_H(\theta_{\pi}) = -\sum_{a=1}^N\sum_{t=0}^{T_{stop}}\sum_{u\in \mathcal{A}} \pi (u_t^a|s_t^a;\theta_{\pi}) \log \pi (u_t^a|s_t^a;\theta_{\pi}).
\end{equation}

Hence, the overall loss for MARL Objective is a combination of the two losses:
\begin{equation}
    \mathcal{L}_{MARL}(\theta_{\pi}) = \mathcal{L}_J(\theta_{\pi}) +\lambda_1 \mathcal{L}_H(\theta_{\pi}),
\end{equation}
where $\lambda_1$ is a constant scaling factor.

\subsubsection{Classification Objective}
A cross-entropy loss is applied to minimize the KL divergence between the ground truth distribution $y$ and prediction $p$:
\begin{equation}
    \mathcal{L}_{Cls}(\theta_p) = - \sum_{c=1}^C y_c\log p_c .
\end{equation}

Finally, we minimize a hybrid loss that combines all the losses:
\begin{equation}
    Loss = \mathcal{L}_{Cls}(\theta_p) + \lambda_2\mathcal{L}_{MARL}(\theta_{\pi}),
\end{equation}
where $\lambda_2$ is a constant scaling factor.

\begin{table*}
\begin{center}
\setlength{\tabcolsep}{4mm}
\begin{tabular}{|c|c|c|c|c|c|c|c|c|}
\hline
\multirow{2}{*}{Architecture} & 
\multicolumn{4}{c|}{ActivityNet v1.2} & 
\multicolumn{4}{c|}{ActivityNet v1.3} \\ 
\cline{2-9}
& R25 & U25 & All & Ours & R25 & U25 & All & Ours \\ \hline\hline
C3D & 62.06 & 62.89 & 63.00 & \textbf{64.13} & 59.73 & 60.68 & 60.83 & \textbf{62.00}\\ \hline\hline
BN-Inception & 78.76  & 80.02  & 80.50 & \textbf{81.99} & 75.08  & 76.48 & 77.33 & \textbf{78.32}\\ 
ResNet-101 & 80.73  & 81.94  & 82.26 & \textbf{83.76} & 78.69  & 79.96  & 80.64 & \textbf{81.54}\\ 
Inception-V3 & 81.90  & 82.66  & 83.25 & \textbf{85.01} & 79.27 & 80.33  & 80.86 & \textbf{82.34}\\ 
ResNet-152 & 82.72  & 83.71  & 84.07 & \textbf{85.70} & 80.69  & 82.08  & 82.53 & \textbf{83.81}\\ \hline        
\end{tabular}
\end{center}
\caption{Performance comparison of different ConvNet architectures on the ActivityNet dataset. For different architectures, randomly sampling 25 frames (R25), uniformly sampling 25 frames (U25), using all frames (All) and using our method to sample 25 frames (Ours) are evaluated. All architectures are based on ImageNet \cite{deng2009imagenet} pre-trained model, except C3D.}
\label{tab:backbones}
\end{table*}

\section{Experiments}
\subsection{Datasets and Evaluation Metrics}

\textbf{ActivityNet} is a large-scale video benchmark for human activity understanding \cite{caba2015activitynet}. The first version of this dataset (termed as ActivityNet v1.2) has 100 classes of human activities.
and its second version (termed as ActivityNet v1.3) contains 19,994 videos from 200 activity categories. 
Moreover, the ActivityNet dataset has temporal annotations of action instances for training data. It is also worth noting that the labels of the test set are not publicly available and thus the performances on ActivityNet dataset are all reported on the validation set. We decode videos at 1fps and use RGB frames in our experiments. Following the official evaluation script, the \textbf{evaluation metric} is based on mean average precision (mAP) for action recognition on ActivityNet.  

\textbf{YouTube Birds and YouTube Cars} are two challenging video datasets for fine-grained video classification which consist of 200 different bird species and 196 different car models respectively \cite{zhu2018fine}. 
We experimented with the RGB frames of the two datasets. Videos in YouTube Birds and YouTube Cars were downsampled to 2fps and 4fps respectively. We employ top-1 precision as the \textbf{evaluation metric} for the two datasets.

\subsection{Implementation Details}
One of the major differences in current video architectures is whether the convolutional operators use 2D (image-based) or 3D (video-based) kernels. Hence, we choose two successful video classification methods for feature extraction in our method, namely temporal segment network \cite{tsn} and C3D \cite{c3d}. The temporal segment network is equipped with segmental modeling (5 segments) to capture long-range temporal information and C3D uses 3D ConvNets to extract the temporal and spatial information of a video clip.

\textbf{Training} When using 2D ConvNets as our backbone, we first pre-train classification network $f_p$ with the method introduced in \cite{tsn}. For 3D ConvNets, we use the C3D \cite{c3d} features provided by ActivityNet's \cite{caba2015activitynet} website which are extracted every 8 frames with a temporal resolution of 16 frames. 
We use the pre-trained weights to initialize $f_p$, then jointly train the network with extra components. Adam with the initial learning rate of 0.0001 is adopted. The parameters used in the experiments are set as follows. We set $F$ to 120, 100 and 100 for ActivityNet, YouTube Birds and YouTube Cars respectively. For the videos shorter than $F$ frames, we cyclically repeat their frames to derive a video of $F$ frames. In the policy network, we use a gated recurrent unit (GRU) cell with the hidden size of 1024 to model the sequential decision process and $T_{max}$ is set to 10. $\gamma$ is empirically set to 0.9 and $\lambda_1, \lambda_2$ are set to 1. We set $N$ to 5 during training.

\textbf{Testing} Many state-of-the-art methods rely on some sophisticated testing strategy or post-processing techniques, such as 10-crop testing, to boost performance. However, our framework does not demand these testing strategies. We simply sample 25 frames ($N$ is set to 25) uniformly per video as the initial temporal location, and single-center cropping is applied to the selected frames from MARL to make the final prediction directly. During evaluation, we use maximum a posterior estimation to choose the action for each agent according to $\pi (u_t^a|s_t^a;\theta_{\pi})$.

\subsection{Improvements over 2D/3D CNN Baselines}
\textbf{MARL improves various backbones}  
Here we examine our framework with several recent network architectures on the ActivityNet dataset using the RGB modality. Specifically, we equip MARL to five deep architectures: BN-Inception \cite{bn-inception}, Inception-V3 \cite{inception-v3}, ResNet-101 \cite{resnet}, ResNet-152 \cite{resnet}, and C3D \cite{c3d}. For each architecture, we evaluate three hand-crafted frame sampling strategies as baselines: randomly sampling 25 frames (R25), uniformly sampling (U25) and using all frames (All). For random sampling, we evaluate three times and report the average results. Since the average duration of videos in ActivityNet is 117s and we decode videos at one fps, we set $F$ to be 120, which indicates that using all frames means using 120 frames. For 2D ConvNets, we follow the TSN \cite{tsn} framework and train the network on annotated action instances, all these 2D ConvNets are initialized with ImageNet pre-trained weights \cite{deng2009imagenet}. For C3D, we also carry out experiments with TSN-style training strategy by sampling 5 clips and predicting labels based on consensus. An average pooling on logits predicted from each sampled frame/clip is followed to produce video-level predictions for different sampling strategies in these experiments.

The results are summarized in Table~\ref{tab:backbones}. We observe that deeper models with higher accuracy (on ImageNet dataset~\cite{deng2009imagenet}) result in better performance in the video classification task, and our method consistently obtains robust improvements over various 2D/3D architectures. For each architecture, our method achieves the best performance and improved over them by a large margin on both datasets comparing with randomly sampling, uniformly sampling and using all frames. 

We can see that even with a very powerful ResNet-152 backbone, MARL can largely boost recognition performance, specifically, the gain over U25 and All is 1.99\% and 1.63\% on ActivityNet v1.2, and as is 1.73\% and 1.28\% on ActivityNet v1.3. This verifies the effectiveness of our method, regardless of shallower or deeper baseline models.

\begin{table}
\begin{center}
\begin{tabular}{|c|c|c|}
\hline
Strategies & Instance & Video \\
\hline\hline
R25/U25/All & 80.69/82.08/82.53 & 80.17/81.23/81.73 \\
\hline
Ours & \textbf{83.81} & \textbf{82.98} \\
\hline
\end{tabular}
\end{center}
\caption{Performance comparison between different supervision on ActivityNet v1.3 val set.}
\label{tab:instance}
\end{table}

\textbf{Instance level v.s. video level supervision}
Annotations of action instances in untrimmed videos are expensive and hardly available under most circumstances. Experiments are carried out to show that our method consistently improves over the baseline method. Besides using these annotated activity instances when training the network, we train our model when only the video-level labels are used. In this experiment, the powerful ResNet-152 is utilized as our backbone network. We present the experimental results of our proposed MARL with annotated instances information (Instance) and video-level information (Video) in Table~\ref{tab:instance}. From the results of ActivityNet v1.3 val set, we observe that our method achieves the best performance in comparison with the three hand-crafted baseline strategies, whether instance-level supervision is available or not. It is also worthy of noting that, our method with video-level information achieves even better performance than the model trained with instance-level supervision and tested with all frames. It further confirms that MARL is effective in picking discriminant frames in untrimmed videos.

\subsection{Comparison with State-of-the-arts}

\begin{figure}
\begin{center}
\includegraphics[width=0.9\linewidth]{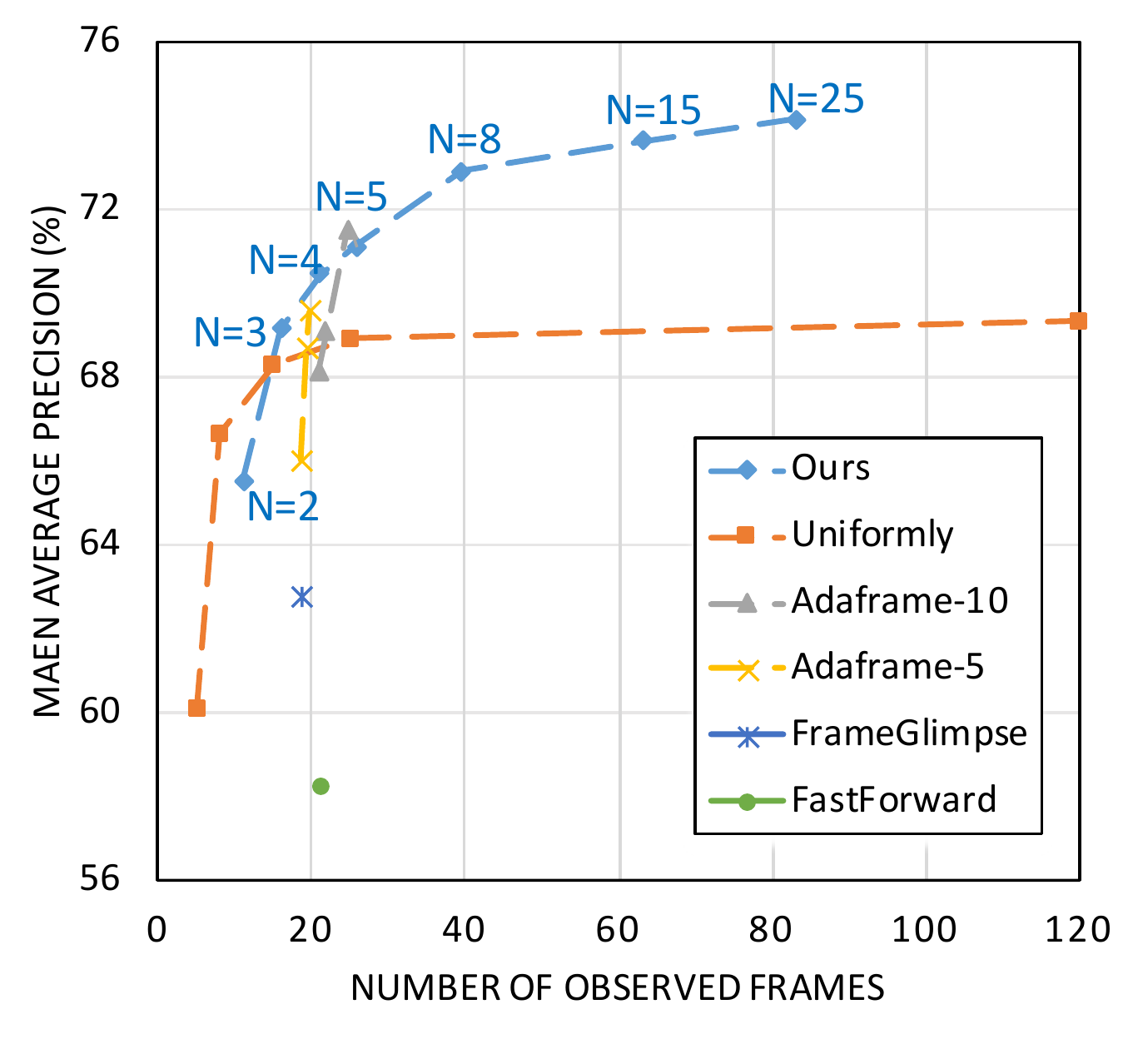}
\end{center}
\caption{ \textbf{Mean average precision vs. number of observed frames.} Comparisons of our method with AdaFrame \cite{wu2018adaframe}, FrameGlimpse \cite{lifeifei_loc}, Fast-Forward \cite{fan2018watching}, and uniformly sampling.}
\label{fig:see_frames}
\end{figure}

\textbf{Comparison with other frame selection methods} We make comparisons under the scenario of observing the same number of total frames and results demonstrate our method outperforms other methods as shown in Figure~\ref{fig:see_frames}. 
The results of these methods are directly cited from Adaframe \cite{wu2018adaframe} whose authors reproduce FrameGlimpse \cite{lifeifei_loc} and Fast-Forward \cite{fan2018watching}. To be fair, we use the same preprocessing method, backbone network (\ie ResNet-101) and training strategy of backbone as Adaframe \cite{wu2018adaframe}. Note that the performances of our solution in Figure~\ref{fig:see_frames} are lower than those reported in Table \ref{tab:backbones}, this is because the backbone is not trained as what TSN \cite{tsn} does for a fair comparison. In this experiment, the number of observed frames is calculated by averaging the total number of frames observed for each video at all steps of MARL.

\begin{table}
\begin{center}
\small
\begin{tabular}{|c|c|c|c|c|c|}
\hline
Method &Backbone &Pre-trained & top-1 & mAP\\
\hline\hline
IDT \cite{wang2013action} &- & ImageNet & 64.70  & 68.69 \\
C3D \cite{p3d} &- & Sports1M & 65.80 & 67.68  \\
TSN \cite{tsn}$^*$ &BN-Inception & ImageNet & 72.97 & 76.56 \\
P3D \cite{p3d} &ResNet-152 & ImageNet & 75.12 & 78.86  \\
RRA \cite{zhu2018fine} &ResNet-152 & ImageNet & 78.81 & 83.42 \\ 
Ours &ResNet-152 & ImageNet & \textbf{79.82} & \textbf{83.81} \\ \hline\hline
TSN  \cite{tsn}$^*$ & BN-Inception & Kinetics & 78.98 & 81.80\\ 
Ours &BN-Inception & Kinetics &  80.22 & 83.52 \\ 
C16 &Ensemble & - &- & \textbf{90.9} \\
Ours &SEResNeXt152 & Kinetics & \textbf{85.72} & 90.05 \\

\hline
\end{tabular}
\end{center}
\caption{Comparing with methods on the ActivityNet v1.3 validation dataset using RGB modality. * indicates the results of our implementation. C16 denotes the champion submission of the ActivityNet 2016 challenge, it fuses multiple powerful models and multi-modal (RGB, optical flow and audio) results.}
\label{tab:anetv1.3}
\end{table}

\begin{table}
\begin{center}
\small
\begin{tabular}{|c|c|c|}
\hline
Method &  YouTube Birds & YouTube Cars\\
\hline\hline
BN-Inception$^*$ & 60.13 & 61.96 \\
I3D \cite{i3d}$^*$ & 40.68 & 40.92\\
TSN \cite{tsn}$^*$ & 72.361 & 74.340\\ 
RRA \cite{zhu2018fine}$^*$ & 73.205 & 77.625\\ \hline\hline
U25/All/Ours & 76.56/76.77/\textbf{79.01} & 76.49/76.99/\textbf{79.77} \\
\hline
\end{tabular}
\end{center}
\caption{Comparing with methods on YouTube Birds and YouTube Cars. * indicates the results of the method come from the latest project page of these datasets.}
\label{tab:yt-dataset}
\end{table}

\begin{figure*}[!t]
\begin{center}
\includegraphics[width=0.9\linewidth]{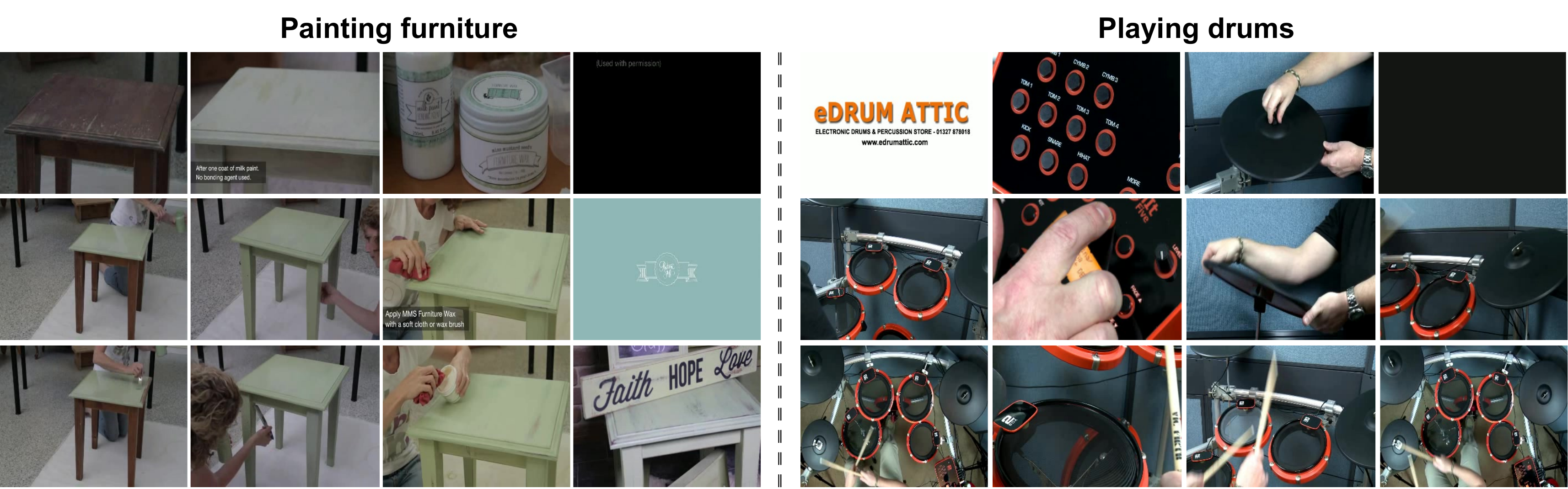}
\end{center}
\caption{Visualization of the selected frame with different strategies on ActivityNet. The first row show frames from uniformly sampling, the second row depicts frames from our method without context-aware observation while the last row contains frames from our method.}
\label{fig:vis}
\end{figure*}

\textbf{Comparison with other SOTAs} We also compare our method with other state-of-the-art methods on these challenging datasets. Table~\ref{tab:anetv1.3} shows results on ActivityNet v1.3, where the results of state-of-the-art-methods all come from published paper or tech reports. We compare with several well known video recognition methods, which once achieved the state-of-the-art performance, including improved trajectories (iDT) \cite{wang2013action}, 3D convolutional networks (C3D) \cite{c3d}, temporal segment networks (TSN) \cite{tsn}, Pseudo-3d residual network (P3D) \cite{p3d}, and Redundancy Reduction Attention (RRA) \cite{zhu2018fine}. We see that our method with ResNet-152 outperforms all these previous methods on the ActivityNet v1.3. 
Moreover, as shown in Table~\ref{tab:anetv1.3}, it is beneficial to pre-train a model using Kinetics dataset\cite{kay2017kinetics} and then transfer it to the video classification task on ActivityNet dataset. ``SEResNeXt152'' pre-trained on Kinetics and finetuned on ActivityNet1.3 is the current SOTA model which won the champion of Action Proposal task in ActivityNet2018\cite{ghanem2018activitynet}. We conducted experiments with RGB modality based on ``SEResNeXt152'', it achieves SOTA mAP of 87.95\% (U25), 88.15\% (All) and 90.05\% (MARL). For readers' reference, here we list the results of the 2016 champion submission as well as ours with SE-ResNeXt-152 \cite{senet} pre-trained on Kinetics dataset. To be noted, our results on ActivityNet v1.3 are obtained by only using 1fps RGB frames during the training phase and a single-center cropping strategy in validation. Even with this simple train and test setup, our single RGB model still achieves comparable results with the 2016 champion submission which fuses multiple powerful models and integrates RGB, optical flow and audio modalities.

Table~\ref{tab:yt-dataset} shows results on YouTube Birds and YouTube Cars. To make a fair comparison, the Inception-V3 \cite{inception-v3} with the same architecture as RRA \cite{zhu2018fine} is used as the backbone. Our method surpasses RRA on all these two datasets since categories in fine-grained tasks often share a similar appearance in general and hence require to focus more on the informative frames to distinguish from each other. Our best performance is 5.805\% above that of other methods on the YouTube Birds and 2.145\% on the YouTube Cars.

\subsection{Ablation Study} 
\textbf{Number of agents} Our model is trained with $N$ of 5 and we evaluate how the testing performance of ActivityNet 1.3 val set varies when $N$ changes. Results can be found in Table~\ref{tab:n}. It can be observed when $N$ increases, the performance gradually improves at first. There is a very interesting phenomenon. When $N$ reaches the total number of frames (which is 120), the performance slightly drops to 83.72\% but is still better than feeding all frames directly (82.53\%). It is because that some agents might select the same frame to avoid less informative frames, but not all the irrelevant frames are skipped by the agents. In this case, though performance drops, it is still better than 82.53\%. Such results evident the variation of frame-level saliency in the untrimmed video and support our motivation. 

\begin{table}
\begin{center}
\begin{tabular}{c|c c c c|c}\hline
  &N=5 &N=15 &N=25 &N=120 &All\\
 \hline
 ResNet-152 & 80.19 & 82.99 & 83.81 & 83.72 &82.53\\
 \hline
\end{tabular}
\end{center}
\caption{Impact of $N$ on ActivityNet v1.3 val set in terms of mAP.}
\label{tab:n}
\end{table}

\textbf{Context range}
We explore the impact of parameter $M$ in context-aware observation network by ablation study. We carry out experiments with MARL frameworks with various settings of $M$. The experimental results are listed in Table~\ref{tab:context}. Basically, our method still works for untrimmed videos when $M=0$. It outperforms uniformly sampling by 0.91\% and is even better than using all frames. The use of the context-aware observation network improves the non-context model with clear margins. However, no obvious gain is obtained by increasing $M$. Larger $M$ means larger network size and cost, so we empirically set $M$ to 1.

\textbf{Policy network transferring} We show that the learned policy network can still be effective when it is transferred directly.
(1) With Resnet-152, performances of directly applying sampling networks trained on ActivityNet1.2/Youtube-Cars(Birds) to ActivityNet1.3 are shown in Table \ref{tab: cross}, see Cars'S, Birds'S and ANet1.2'S. (2) Our policy network trained for ResNet-101 classifier still works for ResNet-152 classifier on ActivityNet1.3, which is indicated as ``cross-cls'' in Table \ref{tab: cross}.

\begin{table}[h]
\begin{center}
\small
\begin{tabular}{c|c|c|c|c|c|c} \hline
    & U25 &All &M=0 &M=1 &M=2 &M=4 \\ \hline
mAP & 82.08 & 82.53 & 82.99  &83.81 & 83.80 & 83.72  \\ \hline
\end{tabular}
\end{center}
\caption{Evaluation of context range on ActivityNet v1.3 val set using ResNet-152. U25 and All are hand-crafted strategies.}
\label{tab:context}
\end{table}

\begin{table}[h]
\scriptsize{
\begin{center}
\begin{tabular}{c c c|c c c |c} \hline
 U25 &All &MARL &Birds'S &Cars'S &ANet1.2'S &cross-cls \\ \hline
82.08 & 82.53 &83.81 &82.70 &82.66 & 83.41 &83.43 \\ \hline
\end{tabular}
\end{center}
}
\caption{The performance of different settings on ActivityNet v1.3 val set.}
\label{tab: cross}
\end{table}

\textbf{Qualitative results} 
We also visualize some examples of selected frames with different strategies on the validation data of ActivityNet in Figure~\ref{fig:vis}. From top to bottom, frames picked by uniformly sampling, $M=0$ and $M=1$ are depicted. We see that our method is able to automatically select important frames according to surroundings and to avoid irrelevant frames.

\section{Conclusion}
In this paper, we presented a multi-agent reinforcement learning method for untrimmed video recognition, which can be effectively applied to existing video recognition frameworks to select the informative frames/clips from the untrimmed video. Experiments demonstrate that our method outperforms state-of-the-art baseline methods by a substantial margin, which verifies the effectiveness of our method. In the future, we plan to distillate our MARL based recurrent frame sampling network into a smaller feed-forward CNN to achieve more efficient untrimmed video classification.   

\section*{Acknowledgment}
The work was funded by National Natural Science Foundation of China (U1713203) and Shenzhen Science and Technology Innovation Commission (Project KQJSCX20180330170238897).

{\small
\bibliographystyle{ieee}
\bibliography{reference}
}

\end{document}